\documentclass[runningheads]{llncs}
\usepackage{graphicx}
\usepackage{amsmath,amssymb} % define this before the line numbering.
\usepackage{color}
\usepackage{epsfig}
\usepackage{graphicx}
\usepackage{amsmath}
\usepackage{amssymb}
\usepackage{breqn}

\begin{document}

\title{Learning Deep Representations with Probabilistic Knowledge Transfer} 

\titlerunning{}

\author{Nikolaos Passalis \and
Anastasios Tefas}

\authorrunning{N. Passalis and A. Tefas}

\institute{Aristotle University of Thessaloniki, Thessaloniki  541 24, Greece 
\email{passalis@csd.auth.gr, tefas@aiia.csd.auth.gr}
}
\maketitle              % typeset the header of the contribution
\begin{abstract}

Knowledge Transfer (KT) techniques tackle the problem of transferring the knowledge from a large and complex neural network into a smaller and faster one. However, existing KT methods are tailored towards classification tasks and they cannot be used efficiently for other representation learning tasks. In this paper we propose a novel probabilistic knowledge transfer method that works by matching the probability distribution of the data in the feature space instead of their actual representation.  Apart from outperforming existing KT techniques, the proposed method allows for overcoming several of their limitations providing new insight into KT as well as novel KT applications, ranging from KT from handcrafted feature extractors to {cross-modal} KT from the textual modality into the representation extracted from the visual modality of the data.

\end{abstract}

\section{Introduction}
\label{section:intro}

\textit{Deep Learning} (DL) has been used to tackle many difficult problems~\cite{lecun2015deep}, ranging from performing accurate object detection~\cite{redmon2016you}, to tackling challenging information retrieval problems~\cite{nips1}, with great success. However, apart from developing more accurate models, the interest of the scientific community has also shifted into creating smaller and faster models that are able to run on devices with limited processing power, such as mobile phones, robots, embedded systems, etc. Several methods have been proposed to this end, including, but not limited to, model compression~\cite{hubara2016binarized}, and lightweight and more efficient neural network architectures~\cite{Bulat_2017_ICCV,howard2017mobilenets,iandola2016squeezenet,Passalis_2017_ICCV}.

\textit{Knowledge Transfer} (KT) techniques have also been proposed to further improve the performance of lightweight neural networks~\cite{hinton2015distilling,romero2014fitnets}. KT works by transferring the knowledge from a powerful and complex model, called \textit{teacher} model, to a smaller and simpler one, called \textit{student} model. Usually, the knowledge is transferred between the models by having the student model to regress the output (or a transformed version of the output) of the teacher model. KT techniques allow for learning student networks that are more accurate and generalize better since the output of the teacher model implicitly encodes more information about the {similarity} between the training samples and their distribution (which is usually ignored during the training when the hard binary labels of the training set are used). In that way, KT acts as a regularizer that improves the performance of the student model~\cite{tang2016recurrent}. Note that KT methods are complementary to other techniques that allow for deploying smaller and faster networks, e.g., MobileNets that use depth-wise separable convolutions~\cite{howard2017mobilenets}, or binarized networks~\cite{Bulat_2017_ICCV}, and they can be combined with them to further improve the accuracy of the models.

\begin{figure}[t]
	\begin{center}
		\includegraphics[width=0.99\linewidth]{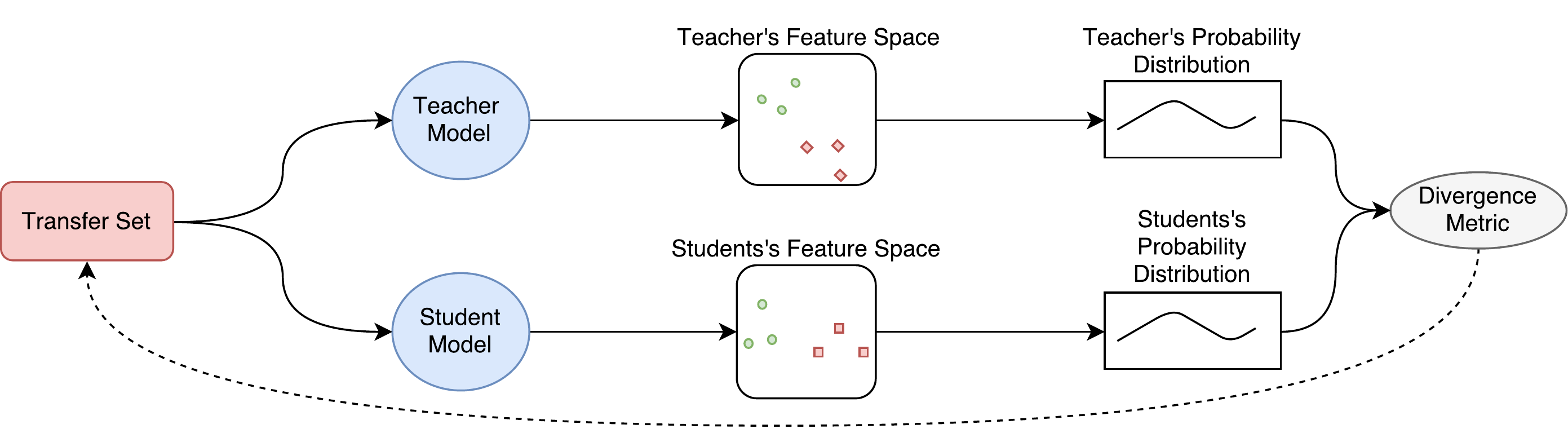}
	\end{center}
	\caption{Probabilistic Knowledge Transfer: The knowledge of the teacher model is modeled using a probability distribution. Then, the knowledge is transferred by minimizing the divergence between the probability distribution of the teacher and the student. }
	\label{fig:short}
\end{figure}

Even though existing KT techniques have been used with great success, they suffer from significant limitations. First, they are usually unable to directly transfer the knowledge between layers of different architecture/dimensionality. The reason for this is that KT methods are currently tailored towards classification tasks, where they are used to transfer the knowledge between the output classification layer of the networks (which has fixed size regardless the actual architecture of the networks). However, this renders most KT methods inappropriate for  \textit{representation learning} tasks that are needed for applications other than classification, such as text and multimedia {information retrieval}~\cite{chitrakar2016social,manning2008introduction}, learning convolutional feature extractors for object detectors~\cite{redmon2016you}/trackers~\cite{nam2016learning}, or {interactive exploratory data analysis}~\cite{fisher2012interactions}. Note that there is a growing demand for learning such lightweight feature extractors, e.g., extracting privacy-preserving
representations (the user's data remain on the mobile device
protecting his/her privacy)~\cite{shokri2015privacy}, lowering the energy and communication costs between mobile devices and the cloud~\cite{cao2015spiking}, etc. Furthermore, existing KT techniques mostly ignore the \textit{geometry} of the teacher's feature space, e.g., manifolds that are formed, similarities between neighboring samples, etc., since they merely regress the output of the teacher network. However, it has been shown that exploiting such kind of information can significantly increase the quality of the learned model regardless the domain of the application~\cite{belkin2006manifold}.

The aforementioned observations led us to a number of interesting questions. a) Is it possible to use existing KT techniques for representation learning tasks instead of merely classification tasks? If this is indeed possible, how does existing KT techniques perform on these tasks?  b) Is there any way to learn a student model
that directly regresses the geometry of the teacher's feature space instead of its output? This could possibly allow to effectively unwrap the manifolds formed in the feature space of the teacher model into the student's lower dimensional feature space increasing the accuracy of the student model. c) Is it possible to transfer the knowledge from handcrafted features, e.g., SIFT~\cite{lowe2004distinctive}, and HoG~\cite{dalal2005histograms}, into a neural network that can be then finetuned towards that task at hand? This could provide a way to exploit the large amount of the available unlabeled training samples and effectively use them in the process of training deep neural networks, overcoming a significant drawback of deep learning models, i.e., the enormous amount of labeled data that are needed for successfully training them. In that way, KT can also significantly boost DL on domains where there is knowledge on how to design good handcrafted representations, but large annotated
collections of data do not exist yet, e.g., high-frequency trading analysis~\cite{kercheval2015modelling}, predicting various properties of chemical compounds~\cite{montavon2013machine}, etc.  d) Finally, can the knowledge of networks which were trained to solve other tasks, such as object detection~\cite{redmon2016you}, be effectively transferred into other smaller networks? This can be especially important since most deep object detectors rely on pre-trained deep convolutional neural networks, while training them from scratch is difficult and usually negatively affects their accuracy~\cite{Shen_2017_ICCV}. Therefore, transferring the knowledge from a larger, pre-trained network into a smaller one can significantly increase the accuracy of lightweight object detectors.

To overcome the limitations of existing KT techniques a probabilistic method for knowledge transfer is proposed in this paper. First, the interactions between the data samples in the feature space are modeled as a \textit{probability distribution} that expresses the affinity between the data samples. In that way, it is possible to perform KT by learning a student model that directly regresses the {probability distribution} of the teacher representation instead of the actual output of the network.  As it is demonstrated in Section~\ref{section:proposed-approach}, this process is connected to an information-theoretic measure, the Mutual Information (MI)~\cite{cover2012elements}, and provides several advantages over existing KT techniques. First, it allows for directly transferring the knowledge even when the output dimensionality of the networks does not match. Furthermore, even when the output dimensionality of the networks matches, directly regressing their outputs might not be the most effective strategy since the teacher network is expected to be less powerful than the student network. Using a method that is able to relax this constraint, e.g., by allowing to slightly distort the feature space, is expected to better facilitate the knowledge transfer process. Finally, note that the probability distribution can be also estimated or enhanced using any other information source, such as neural network ensembles, handcrafted feature extractors, supervised information or even qualitative information from domain experts or users, increasing the flexibility of the proposed method and allowing for using several new KT scenarios.

The main contribution of this paper is the proposal of a Probabilistic KT (PKT) technique that overcomes several limitations of existing KT methods by matching the probability distribution of the data in the feature space instead of their actual representation, as shown in Figure~\ref{fig:short}. To the best of our knowledge the proposed technique is the first that is capable of a) performing cross-modal knowledge transfer, b) transferring the knowledge from handcrafted feature extractors into neural networks, c)	transferring the knowledge regardless the task at hand (e.g., object detection), and d) incorporating domain-knowledge into the knowledge transfer procedure, providing new insight into KT. The proposed method is motivated by the fact that matching the  probability density function of the teacher and student models maintains the teacher's \textit{quadratic mutual information} (QMI)~\cite{torkkola2003feature}, between the feature representation of the data samples and a set of (possible unknown) label annotations. Also, the proposed method is capable of recreating the (local) geometry of the teacher's feature space into the feature space of the student model. Indeed, the proposed method  embeds the manifolds formed in teacher's feature space into the student's space (regardless of the dimensionality of these spaces). The proposed method is extensively evaluated and compared to other KT techniques using four different evaluation setups  (KT from deep neural networks, handcrafted feature extractors, different modalities and object detectors). Also, it is demonstrated that is possible to perform \textit{cross-modal} KT by transferring the knowledge from the textual modality into the representation extracted from the visual modality.   An easy to use implementation of the proposed method is available at \url{https://github.com/passalis/probabilistic\_kt} to allow for easily using and extending the proposed method.

The rest of the paper is structured as follows. The related work is presented and compared to the proposed approach in Section~\ref{section:related-work}. Then, the proposed method is presented in detail in Section~\ref{section:proposed-approach} and evaluated in   Section~\ref{section:experiments}. Finally, conclusions are drawn in Section~\ref{section:conclusions}.

\section{Related Work}
\label{section:related-work}

The growing complexity of deep neural networks and the need to deploy them into mobile and embedded devices with limited computing capabilities have fueled the research on knowledge transfer techniques that are able to effectively train smaller and faster models. The vast majority of the proposed KT methods use the teacher model to generate soft-labels, e.g., by raising the temperature of the softmax activation function on the output layer of the network, that are then used to train the student model \cite{compression-model,chan2015transferring,hinton2015distilling,tang2016recurrent,tzeng2015simultaneous}.

The first attempt for knowledge transfer using soft-labels was presented in \cite{compression-model}, while the neural network distillation method~\cite{hinton2015distilling}, extends this approach by appropriately tuning the temperature of the softmax activation function. It has been demonstrated that the neural network distillation method can be used to efficiently regularize the smaller network and achieve better generalization than directly training the network using the labels of the training set~\cite{chitrakar2016social,hinton2015distilling,howard2017mobilenets}. Furthermore, the generated soft-targets can be used for pre-training a larger network, as in~\cite{tang2015knowledge}, for domain adaptation in combination with sparsely labeled data, as in~\cite{tzeng2015simultaneous}, or  for ``compressing'' the posterior predictive density in Bayesian methods~\cite{balan2015bayesian}. Also, in~\cite{chan2015transferring}, the knowledge is transferred from a recurrent neural network (RNN) to a deep neural network using a similar approach. It is worth mentioning a quite opposite approach followed in~\cite{tang2016recurrent}, where the knowledge is transferred from a weaker teacher model to a more powerful student network. It was demonstrated that this allows for training the student network using fewer labeled data and it highlights the regularization nature of the distillation process. 

The aforementioned methods use soft-labels to train the student network. A vastly different approach is followed in \cite{chen2015net2net}, where the weights of the teacher model are used to initialize the student model allowing for faster convergence. Furthermore, in \cite{romero2014fitnets}, the student network is trained not only using the soft-targets, but also using \textit{hints} from the intermediate layers. Since the size of the student model is usually smaller, this is achieved by using a projection to match the dimensionality of the targets and the output of the student model. A similar approach is also followed in~\cite{yim2017gift}, where instead of using hints, the \textit{flow of solution procedure} (FSP) matrix is used to transfer the knowledge between some of the intermediate layers of a residual network. However, in contrast to the hint-based transfer method, the FSP-based method requires the intermediate layers of the networks to have the same size and number of filters, rendering the method unsuitable for representation learning when the dimensionality of the layers between the two networks is different (which is expected to be the case when learning a smaller network).

To the best of our knowledge the method proposed in this paper is the first probabilistic KT method for representation learning that works by directly matching the probability distribution of the data between the teacher's and the student's feature spaces using an appropriately defined divergence metric. The proposed method is simple and straightforward, without requiring careful domain-specific tuning of any hyper-parameter, such as the softmax temperature~\cite{hinton2015distilling}. As we experimentally demonstrate in Section~\ref{section:experiments}, this allows for directly using the proposed method for a wide range of different KT scenarios. Furthermore, the proposed method is capable of directly transferring the knowledge between spaces of different dimensionality by modeling the interactions between the data samples  and, thus, avoiding the need for lossy low dimensional projections~\cite{romero2014fitnets}.  Also, the proposed method requires no knowledge about the teacher model, except for the probability distribution  induced by the representation of the data samples, significantly increasing its flexibility and allowing for novel KT scenarios, such as  transferring the knowledge from handcrafted feature extractors. This is in contrast with other methods that require having access to the weights of the teacher network~\cite{chen2015net2net}. The  probability distribution can be also enhanced using domain knowledge or supervised information providing a straightforward way to directly incorporate such information into the KT procedure. Finally, the proposed method can be also used for classification tasks, similarly to other methods that regularize the distillation process by transferring the knowledge between intermediate layers, such as~\cite{passalis18,romero2014fitnets,yim2017gift}.

\section{Probabilistic Knowledge Transfer}
\label{section:proposed-approach}

%First, the MI and its quadratic variant are briefly introduced in this Section. Then, the proposed method is derived by learning a teacher model that maintains the same amount of mutual information between the extracted representation and a set of labels as the teacher model. The proposed method is presented in detail and several design choices are discussed through this Section.

Let $\mathcal{T} =\{ \mathbf{t}_1, \mathbf{t}_2, \dots, \mathbf{t}_N \}$ denote a collection of $N$ objects that are used to transfer the knowledge between two models. The set $\mathcal{T}$ is also called \textit{transfer set}.  Also, let $\mathbf{x} =f(\mathbf{t})$ denote the output representation of the teacher model and $\mathbf{y}=g(\mathbf{t}, \mathbf{W})$ denote the output representation of the student model, where $\mathbf{W}$ denotes the parameters of the student model. During the process of knowledge transfer the parameters $\mathbf{W}$ of the model $g(\cdot)$ are learned to ``mimic'' the behavior of $f(\cdot)$. Note that there is no constraint on what the functions $f(\cdot)$ and $g(\cdot)$ are  as long as the output of $f(\cdot)$ is known for every element of $\mathcal{T}$ and $g(\cdot)$ is a  differentiable  function. The distributions of the teacher and student networks are modeled using two continuous random variables ${X}$ and ${Y}$ respectively, where ${X}$ describes the representation extracted from the teacher model and ${Y}$ describes the representation extracted from the student model.

Modeling the pairwise interactions between data samples allows for describing the geometry of the corresponding feature spaces~\cite{hinton2003stochastic,maaten2008visualizing}. To this end, the joint probability density of any two data points in the feature space, which models the probability of two data point being close together, can be used. To this end, the divergence between the joint density probability estimations for the teacher model $\mathcal{P}$ and the student model $\mathcal{Q}$ can be minimized. These joint density probability functions can be trivially estimated using Kernel Density Estimation (KDE)~\cite{scott2015multivariate} as:
\begin{equation}
\label{eq:j1}
p_{ij} = p_{i|j} p_{j} =  \frac{1}{N} K(\mathbf{x}_i,\mathbf{x}_{j}; 2\sigma_t^2), 
\end{equation}
and
\begin{equation}
\label{eq:j2}
q_{ij} = q_{i|j} q_{j} = \frac{1}{N} K(\mathbf{y}_i, \mathbf{y}_{j}; 2\sigma_s^2), 
\end{equation}
respectively, where $K(\mathbf{a}, \mathbf{b};  \sigma_t^2)$ is a symmetric kernel with width $\sigma_t$ and $\mathbf{a}$ and $\mathbf{b}$ are two vectors. Note that class labels are not needed to minimize the divergence between these two distributions. Therefore, the proposed method can be used even when the class labels are unknown. Also note that minimizing the divergence between the probability distribution of the teacher model $\mathcal{P}$ and the probability distribution of the student model $\mathcal{Q}$ ensures that each transfer sample will have the same neighbors in both the student and teacher spaces as well as the relative distances between samples will be maintained. This, in turn, implies that the geometric relationships of the teacher's feature space are maintained in the lower dimensional feature space of the student. 

Using the joint probability distribution to model the {geometry} of the data and perform knowledge transfer can overcome many of the drawbacks of traditional KT methods (as discussed in Section~\ref{section:related-work}). However, learning a significantly smaller model that accurately recreates the whole geometry of a complex teacher model is often impossible. To overcome this issue, the joint probability density function can be replaced with the conditional probability distribution of the samples. Even in both cases the divergence between the probability distributions is minimized when the kernel similarities are equal for both models, using the conditional probability distribution allows for more accurately describing the local regions between the samples (the conditional probability distribution expresses the probability of each sample to select each of its neighbors~\cite{maaten2008visualizing}). The conditional probability distributions have been also used for the same reason in dimensionality reduction techniques that model data distributions in very high dimensions, such as the \mbox{t-SNE} algorithm~\cite{maaten2008visualizing}. The conditional probability distribution for the teacher model is defined as:
\begin{equation}
\label{eq:teacher-prob}
p_{i|j} = \frac{K(\mathbf{x}_{i}, \mathbf{x}_{j}; 2\sigma_t^2)}{\sum_{k=1, k\neq j}^N K(\mathbf{x}_{k}, \mathbf{x}_{j}; 2\sigma_t^2)} \in [0,1],
\end{equation}
while for the student model as:
\begin{equation}
q_{i|j} = \frac{K(\mathbf{y}_{i}, \mathbf{y}_{j}; 2\sigma_t^2)}{\sum_{k=1, k\neq j}^N K(\mathbf{y}_{k}, \mathbf{y}_{j}; 2\sigma_s^2)} \in [0,1].
\end{equation}
The conditional probabilities are bounded to $[0, 1]$ and sum to 1, i.e., $\sum_{i=0, i\neq j}^N p_{i|j} \\= 1$ and $\sum_{i=0, i\neq j}^N q_{i|j} = 1$.

Several choices exist to define the used kernel. Perhaps the most natural choice is the Gaussian kernel: 
\begin{equation}
K_{Gaussian}(\mathbf{a}, \mathbf{b}; \sigma) = exp\left(-\frac{||\mathbf{a} - \mathbf{b}||^2_2}{\sigma}\right),
\end{equation}
 where $||\cdot||_2$ denotes the $l^2$ norm of a vector and $\sigma$ is the scaling factor (width) of the kernel. Using the Gaussian kernel leads to the regular Kernel Density Estimation (KDE) method for estimating the conditional probabilities~\cite{scott2015multivariate}. However, to ensure that a meaningful probability estimation is obtained the width of the kernels must be carefully tuned. This is not a straightforward task, with several heuristics proposed to tackle this problem~\cite{turlach1993bandwidth}. To avoid this issue and derive a method that requires little domain-dependent tuning, a cosine similarity-based affinity metric is used in this work. Therefore, the employed similarity metric is defined as: 
\begin{equation}
 K_{cosine}(\mathbf{a}, \mathbf{b}) = \frac{1}{2}(\frac{\mathbf{a}^T\mathbf{b}}{||\mathbf{a}||_2 ||\mathbf{b}||_2} + 1) \in [0, 1].
\end{equation} Apart of avoiding the need for calculating the bandwidth of the kernel, using the cosine similarity as kernel metric allows for more robust affinity estimations, since it has been demonstrated that the cosine measure usually leads to improved performance over Euclidean measures (especially in high dimensional spaces)~\cite{manning2008introduction,wang2015visual}.

Also, several choices exist for the divergence metric that must be used for training the student model. In this work, the well known Kullback-Leibler (KL) divergence is used to this end:
\begin{equation}
\mathcal{KL}(\mathcal{P}||\mathcal{Q}) = \int_{\mathbf{t}} \mathcal{P}(\mathbf{\mathbf{t}}) \log \frac{\mathcal{P}(\mathbf{t})}{\mathcal{Q}(\mathbf{t})} d\mathbf{t},
\end{equation}
where $\mathcal{P}$ and $\mathcal{Q}$ are the probability distributions of the teacher and student models respectively. Since a finite number of points are used to approximate the distribution $\mathcal{P}$ and $\mathcal{Q}$, the loss function used for training the model is calculated as:
\begin{equation}
\label{eq:divergence-discrete}
\mathcal{L} = \sum_{i=1}^{N} \sum_{j=1, i\neq j}^{N} p_{j|i}  \log \left(\frac{p_{j|i}}{q_{j|i}}\right).
\end{equation}
Note that the KL divergence is not a symmetric distance metric, giving higher weight to minimizing the divergence for neighboring pairs of points instead of distant ones. That means that maintaining the geometry of local neighborhoods is more important, during the optimization process, than recreating the global geometry of the whole feature space of the teacher, providing greater flexibility during the training of the student model. If maintaining the whole geometry of the feature space is equally important, then alternative symmetric divergence metrics, such as the quadratic divergence measure $D_Q(\mathcal{P},\mathcal{Q}) = \int_{\mathbf{x}}(\mathcal{P}(\mathbf{t}) - \mathcal{Q}(\mathbf{t}))^2d\mathbf{t}$, can be used. However, it should be noted that it is often infeasible to achieve this when training a student model with a significantly smaller number of parameters. 

To learn the parameters of the student model $g(\mathbf{t}, \mathbf{W})$ gradient descent is used, i.e., $\Delta \mathbf{W} =  - \eta \frac{\partial \mathcal{L}}{ \partial \mathbf{W}},$ where $\mathbf{W}$ is the matrix with the parameters of the student model. The derivative of the loss function with respect to the parameters of the model can be easily derived as
\begin{equation}
\frac{\partial \mathcal{L}}{ \partial \mathbf{W}} = \sum_{i=1}^N \sum_{j=1, i \neq j}^N \frac{\partial \mathcal{L}} {\partial q_{j|i}}  \sum_{l=1}^N \frac{\partial q_{j|i}} {\partial \mathbf{y}_l} \frac{\mathbf{y}_l}{\partial \mathbf{W}},
\end{equation}
where $\frac{\mathbf{y}_l}{\partial \mathbf{W}}$ is just the derivative of the student's output with respect to its parameters. Instead of using the plain stochastic gradient descent, a recently proposed method for stochastic optimization, the Adam algorithm~\cite{kingma2014adam}, that calculates adaptive learning rates for each parameter of the model, is used for all the experiments conducted in this paper. Furthermore, the conditional probabilities are estimated using only a small batch of the data (64-128 samples) at each iteration, since it is usually intractable to calculate the full kernel matrix for the whole dataset. This process can be  viewed as a Nystr{\"o}m-like approximation of the full similarity matrix~\cite{drineas2005nystrom}, and it was experimentally established that it speeds up the training process, while it does not negatively impact the learned representation. The transfer samples are shuffled between the training epochs to ensure that different samples are used for estimating the conditional probability distributions during each epoch.

\paragraph{PKT and Mutual Information}
In the following we provide a connection between the proposed method and maintaining the same amount of mutual information (MI) between the learned representation and a set of (possible unknown) labels as the teacher model. MI is a measure of dependence between random variables~\cite{cover2012elements}. Let ${C}$ be a discrete random variable that describes an attribute of the samples, e.g., their labels. For each feature vector $\mathbf{x}$ drawn from $X$ there is an associated label $c$. The mutual information measures how much the uncertainty for the class label $c$ is reduced after  observing the feature vector $\mathbf{x}$~\cite{torkkola2003feature}. Let $p(c)$ be the probability of observing the class label $c$. Also, let $p(\mathbf{x}, c)$ denote the probability density function of the corresponding joint distribution. Then, the mutual information for the teacher is defined as:
\begin{equation}
I(X,C)=\sum_c \int_{\mathbf{x}} p(\mathbf{x}, c) \log \frac{p(\mathbf{x}, c)}{p(\mathbf{x})P(c)}d\mathbf{x}
\end{equation}

MI can be also expressed as the KL divergence between the joint probability density $p(\mathbf{x}, c)$ and the product of marginal probabilities  $p(\mathbf{x})$ and $P(c)$. The Quadratic Mutual Information (QMI) is derived by replacing the KL divergence by a quadratic divergence measure, as proposed in~\cite{torkkola2003feature}: $ I_T(X, C) = \sum_{c} \int_{\mathbf{x}} \left(p(\mathbf{x}, c) - p(\mathbf{x})P(c)\right)^2 d\mathbf{X}$. By expanding this definition, we obtain:
\begin{dmath}
	I_T(X, C)  = \sum_{c} \int_{\mathbf{x}} p(\mathbf{x}, c)^2 d\mathbf{x} + \sum_{c} \int_{\mathbf{x}} (p(\mathbf{x})P(c))^2 d\mathbf{x} -2\sum_{c} \int_{\mathbf{x}} p(\mathbf{x}, c)p(\mathbf{x})P(c) d\mathbf{x},
\end{dmath}
where the following quantities, called \textit{information potentials} of the teacher model, can be defined: $
V^{(t)}_{IN} = \sum_{c} \int_{\mathbf{x}} p(\mathbf{x}, c)^2 d\mathbf{x},$
$	V^{(t)}_{ALL} = \sum_{c} \int_{\mathbf{x}} (p(\mathbf{x})P(c))^2 d\mathbf{x},$
and 
$	V^{(t)}_{BTW}=\sum_{c} \int_{\mathbf{x}} p(\mathbf{x}, c)p(\mathbf{x})P(c) d\mathbf{x}.$
Thus, QMI can be expressed in terms of these information potentials as: $
I_T(X, C)  = V^{(t)}_{IN} + V^{(t)}_{ALL} - 2V^{(t)}_{BTW}$.
Assuming that $N_C$ different (and possible unknown) classes exist and each class is composed of $J_p$ samples, the class prior probability for the $c_p$ class is calculated as $P(c_p)=\frac{J_p}{N}$, where $N$ is the total number of samples used to estimate the QMI. Also, Kernel Density Estimation~\cite{scott2015multivariate}, can be used to estimate the joint density probability as $	p(\mathbf{x}, c_p) = \frac{1}{N} \sum_{j=1}^{J_p} K(\mathbf{x}, \mathbf{x}_{pj}; \sigma_t^2)$, where the notation $\mathbf{x}_{pj}$ is used to refer to the $j$-th sample of the $p$-th class, as well as  probability density of $X$ as $p(\mathbf{x})= \sum_{p=1}^{J_p} p(\mathbf{x}, c_p)= \frac{1}{N} \sum_{j=1}^{N} K(\mathbf{x}, \mathbf{x}_{j}; \sigma_t^2)$.

The information potentials for the teacher model are derived using these probabilities~\cite{torkkola2003feature}:
\begin{equation}
\label{eq:pot1}
V^{(t)}_{IN} = \frac{1}{N^2} \sum_{p=1}^{N_c} \sum_{k=1}^{J_p} \sum_{l=1}^{J_p} K(\mathbf{x}_{pk}, \mathbf{x}_{pl}; 2\sigma_t^2),
\end{equation}
\begin{equation}
\label{eq:pot2}
V^{(t)}_{ALL} = \frac{1}{N^2} \left(\sum_{p=1}^{N_c} (\frac{J_p}{N})^2\right)  \sum_{k=1}^{N} \sum_{l=1}^{N} K(\mathbf{x}_{k}, \mathbf{x}_{l}; 2\sigma_t^2),
\end{equation}
and
\begin{equation}
\label{eq:pot3}
V^{(t)}_{BTW} = \frac{1}{N^2} \sum_{p=1}^{N_c} \frac{J_p}{N} \sum_{j=1}^{J_p} \sum_{k=1}^{N} K(\mathbf{x}_{pj}, \mathbf{x}_{k}; 2\sigma_t^2).
\end{equation}

The interaction between two samples $i$ and $j$ is measured using the kernel function $K(\mathbf{x}_i, \mathbf{x}_j; \sigma^2)$ that expresses the similarity between them. Also, note that all the information potentials are expressed  in terms of interactions between all the pairs of the data (weighted by a different factor). The potential $V_{IN}$ expresses the in-class interactions, the potential $V_{ALL}$ the interactions between all the samples, while the potential $V_{BTW}$ the interaction of each class against all the other samples. Similarly, the information potentials can be calculated for the student network, e.g., $V^{(s)}_{IN} = \frac{1}{N^2} \sum_{p=1}^{N_c} \sum_{k=1}^{J_p} \sum_{l=1}^{J_p} K(\mathbf{y}_{pk}, \mathbf{y}_{pl};  2\sigma_s^2)$. Different (and appropriately tuned) widths $\sigma_t$ and $\sigma_s$ must be used for the teacher and the student model.

If QMI is to be transferred between the models, then this implies that the respective information potentials must be equal between the two models. To have equal information potentials between the two models the values provided by the kernel function for each pair of data samples must be equal, i.e., $K(\mathbf{x}_{i}, \mathbf{x}_{j}; 2\sigma_t^2) = K(\mathbf{y}_{i}, \mathbf{y}_{j}; 2\sigma_s^2)\ \forall i, j$, which in turn implies that the joint densities defined in~(\ref{eq:j1})  and~(\ref{eq:j2}) must be equal to each other.

\section{Experimental Evaluation}
\label{section:experiments}

\textbf{KT from Deep Neural Networks:} First, the proposed method was evaluated using the \mbox{CIFAR10}. The knowledge was transferred from the penultimate layer of a deep neural network, the \mbox{ResNet-18} network~\cite{he2016deep}, that has over 11 million parameters, to a significantly smaller student network with the following architecture: $3\times 3$ convolution with 8 filters, $2\times 2$ max pooling, $3\times 3$ convolution with 16 filters, $2\times 2$ max pooling, $3\times 3$ convolution with 32 filters, $2\times 2$ max pooling and a fully connected layer with $64$ neurons. Batch normalization was used after each convolutional layer~\cite{ioffe2015batch}, and the ReLU activation function was used for all the layers. The student network is composed of approximately 15,000 trainable parameters, i.e., more than 700 times less than the teacher ResNet model. The teacher network was trained for classifying the images of the CIFAR10 dataset (after adding a final classification layer with the softmax activation function) for 100 epochs with a learning rate of 0.001 for the first 50 epochs and a learning rate of 0.0001 for the last 50 epochs. A baseline teacher model was also trained and evaluated using the same setup.

The experimental results are reported in Table~\ref{table:cifar10-results}.  All the methods were evaluated in a content-based retrieval setup, where the database is composed of the representation extracted from training images using the student network $g(\cdot)$, while the test set is used to query the database and evaluate each method. To evaluate the quality of the learned representation the  (interpolated) mean Average Precision   (mAP) at the standard 11-recall points and the top-k precision (abbreviated as ``t-k'') were used~\cite{manning2008introduction}. The cosine similarity was used to measure the similarity between the query and the database objects for all the conducted experiments. The penultimate layers (64-dimensional for the student model $g(\cdot)$ and 512-dimensional for the teacher model $f(\cdot)$) were used to extract the representation of the images and transfer the knowledge. The proposed method was compared to the hint-based knowledge transfer~\cite{romero2014fitnets}, abbreviated as ``Hint'', that supports  directly transferring the knowledge between layers of different dimensionality (only the ``hint'' part of the method was used by employing random projections, since it is not possible to use the distillation approach between layers of different dimensionality). 	Note that neither the distillation approach~\cite{hinton2015distilling}, or the FSP transfer~\cite{yim2017gift}, can be employed when the dimensionality of the layers that are used for the knowledge transfer does not match~\cite{yim2017gift}. To ensure a fair comparison between the evaluated KT methods, the baseline student network was used for initializing the network for all the methods and the optimization process ran for 20 epoch with batch size 128 and learning rate 0.0001. The proposed method was also compared to the plain distillation approach (abbreviated as ``Distill.''), where the knowledge was transferred between the classification layers of the networks. However, it should be noted that this requires adding one extra classification layer to the student network and restricts the number of scenarios where the knowledge transfer can be used (since the knowledge must be transferred from a model that has been trained for classification tasks). Finally, the training data of the dataset were used as the transfer set (without using the supplied class labels).

Several conclusions can be drawn from the results reported in Table~\ref{table:cifar10-results}. First, it is confirmed that the proposed PKT method can indeed lead to significantly better results than directly training the network with the available hard labels of the training set. The hint-based method is capable of transferring some knowledge between the layers, but since it is based on random projections its power is severely limited in this application. As a result, the hint-based transfer actually decreases the retrieval precision. This phenomenon can be better understood if the regularization nature of the hint-based approach is considered, i.e., the hints were intended to regularize distillation process instead of being used for directly transferring the knowledge~\cite{romero2014fitnets}.  On the other hand, the proposed PKT method is capable of efficiently transferring the knowledge, significantly improving the mAP (51.19\%) over the rest of the evaluated methods (40.13\% for the next best performing method). Furthermore, note that the distillation approach cannot be used when the networks are not trained for classification and the dimensionality of the layers used for the transfer does match, while the proposed method can effectively overcome these limitations, as further demonstrated in the following experiments.

\begin{table}[!tb]
	\begin{minipage}{.5\linewidth}
			\centering		
			\caption{CIFAR10 Evaluation}
			\label{table:cifar10-results}	
			\begin{tabular}{l|ccccc}
				\textbf{Model} & \textbf{mAP} &  \textbf{t-10} &  \textbf{t-20} & \textbf{t-50} & \textbf{t-100} \\
				\hline
				
				Student & ${38.96}$ & $68.30$& ${65.35}$ & ${61.89}$ & ${59.17}$  \\
				Teacher &  ${91.39}$ & $93.34$ & ${93.19}$ & ${92.28}$ & ${92.81}$  \\
				\hline
				\hline
				Distill. &  ${40.13}$ & $68.81$ & ${65.95}$ & ${62.55}$ & ${59.93}$  \\
				Hint &  ${21.40}$ & ${33.20}$ & ${30.10}$ & ${27.16}$ & ${24.89}$  \\
				PKT &  $\mathbf{51.19}$ & $\mathbf{69.41}$ & $\mathbf{67.38}$ & $\mathbf{65.10}$ & $\mathbf{63.39}$  \\
			\end{tabular}	
	\end{minipage}%
	\begin{minipage}{.5\linewidth}
			\caption{YouTube Faces Evaluation}
%%			\raggedleftmini			
			\label{table:yt-results}	
			\centering
			\begin{tabular}{l|cccccc}
				\textbf{Model}&  \textbf{mAP}   &   \textbf{t-20} &   \textbf{t-50} & \textbf{t-200} \\
				\hline
				LBP & ${46.38} \pm {0.88}$ & ${98.78} $  & ${95.66} $ & ${81.02}$ \\
				\hline
				\hline
				Hint  & ${52.31} \pm {1.31}$ &  ${98.23}$  & ${96.37}$ & $86.10$  \\
				PKT  & $\mathbf{54.84} \pm \mathbf{0.76}$ &  $\mathbf{99.85}  $ &  $\mathbf{98.95} $ &$\mathbf{88.71}$  \\
				\hline
				S-PKT  & ${70.11} \pm {0.95}$  & ${99.88} $ & ${99.31} $ & $91.50$  \\
			\end{tabular}	
	\end{minipage} 
\end{table}

\textbf {KT from  Handcrafted Feature Extractors:} Next, the proposed method was evaluated using the large-scale YouTube Faces dataset~\cite{wolf2011face}.
Before feeding each face image into the used networks it was resized to $64 \times 64$ pixels.  An evaluation
strategy, similar to those of celebrity face image retrieval
tasks~\cite{guo2016ms}, is used. The persons that appear in more than 500 frames, i.e., 225 persons, are considered popular (celebrities). A total of 260,108 frames were extracted, where the 200,000 of them were used for training the method and building the database and 1,000 of them were used to query the database and evaluate the performance of the methods. The training/evaluation process was repeated five times and the mean and standard deviation of the evaluated metrics are reported.

A completely different evaluation setup was used for this dataset. Instead of transferring the knowledge from a larger neural network, the knowledge was transferred from a handcrafted feature extractor to a small neural network that also reduces the size of the extracted representation. More specifically, the representation extracted using the CS-LBP descriptors was used to perform the KT~\cite{wolf2011face}.  The dimensionality of the CS-LBP descriptors is 480, while the last layer of the network used to perform the knowledge transfer has only 64 neurons. Note that the distillation approach cannot be used in this case since there is no common classification layer and the dimensionality of the extracted representation differs. The following architecture was used for the student network: $3\times 3$ convolution with 16 filters, $3\times 3$ convolution with 32 filters, $2\times 2$ max pooling, $3\times 3$ convolution with 64 filters, $3\times 3$ convolution with 64 filters, $2\times 2$ max pooling, $3\times 3$ convolution with 64 filters and a fully connected layer with $64$ neurons. Again, batch normalization was used after each convolutional layer and the ReLU activation function was used for all the layers. All the knowledge transfer models were trained for 10 epochs using learning rate $0.0001$ and batch size 128.

The evaluation results are reported in Table~\ref{table:yt-results}. The baseline model (LBP features) achieves a mAP of 46.38\%, while the hint-based transfer increases the mAP to 52.31\%. The proposed PKT again outperforms both the baseline and hint-based methods (54.84\% mAP). Note that it is generally not expected the student models to achieve higher precision than the teacher model (the handcrafted feature extractor in our case). This behavior can be attributed to the completely different nature of the teacher model (convolutional neural network instead of a handcrafted feature extractor) in combination with the smaller dimensionality of the extracted representation that effectively regularize the learned representation.

Transferring the knowledge from a handcrafted feature extractor into a neural network also allows for finetuning the learned representation using supervised information (or any other kind of domain-specific knowledge). The proposed PKT method readily supports augmenting the transfer procedure with supervised information by simply constructing an appropriate probability distribution function. The most straightforward way to do so it to set $p_{i|j} = 1$ when the $i$-th and the $j$-th sample belong to the same class and $p_{i|j} = 0$ in any other case. This supervised probability distribution function can be combined with the probability distribution extracted from the representation of the teacher model or used standalone as a separate term in the loss function (by adding another divergence loss in Eq.~(\ref{eq:divergence-discrete})). The latter choice is followed in the proposed Supervised PKT (abbreviated as ``S-PKT'') with the supervised divergence weighted by 0.001. As it is demonstrated in Table~\ref{table:yt-results} the S-PKT significantly increases the evaluated metrics over the unsupervised methods.

\textbf{Cross-modal KT:}
The proposed method was also evaluated on the SUN Attribute dataset~\cite{patterson2012sun}, using an evaluation setup similar to this used for the Youtube Faces dataset, i.e., the knowledge is transferred from a handcrafted feature extractor ($2 \times 2$ HoG features~\cite{dalal2005histograms,patterson2012sun}). The SUN Attribute dataset is a large-scale scene attribute dataset that contains more than 700 categories of scenes and 14,000 images. Each image is also described by 102 discriminative textual attributes that were created by performing crowd-sourced human studies. The confidence for each textual attribute is provided. The multi-modal nature of the dataset makes it very appropriate for evaluating cross-modal and multi-modal techniques~\cite{ozdemir2014probabilistic}. The SUN Attribute dataset was also used in this paper to evaluate the performance of cross-modal KT. Since a very small number of images exist for some categories, only images from the eight most common categories (for which at least 40 images exist) were used for training and evaluating the methods. Each image was resized in $128\times 128$ pixels. The 80\% of the extracted images was used for training the networks and building the database, while the rest 20\% was used to query the database. The evaluation process was repeated 5 times and the mean and standard deviation of the evaluated metrics are reported.

The teacher network is also similar to the network used for the Youtube Faces dataset and it is composed of the following layers: $3\times 3$ convolution with 16 filters, $2\times 2$ max pooling, $3\times 3$ convolution with 32 filters, $2\times 2$ max pooling, $3\times 3$ convolution with 64 filters, $3\times 3$ convolution with 64 filters, $2\times 2$ max pooling, $3\times 3$ convolution with 64 filters and a fully connected layer with $64$ neurons. All the knowledge transfer models were trained for 100 epochs using learning rate $0.0001$ and batch size 128.

\begin{table}[!tb]
	\begin{minipage}{.7\linewidth}
		\centering
			\caption{SUN Attribute Evaluation}
			\label{table:sun-results}			
				\begin{tabular}{l|ccccc}
					\textbf{Model}&  \textbf{mAP} &   \textbf{t-20} & \textbf{t-50}   \\
					\hline
					HoG & ${32.06} \pm {1.20}$  & ${42.65} \pm {2.00}$ & ${34.13} \pm {1.64}$  \\
					\hline
					\hline
					Hint (HoG) & ${16.40} \pm {0.58}$  & ${19.04} \pm {1.07}$ & ${16.85} \pm {0.86}$  \\
					PKT (HoG) & $\textbf{26.84} \pm \textbf{1.74}$ & $\textbf{34.98} \pm \textbf{2.42}$ & $\textbf{28.96} \pm \textbf{1.98}$  \\
					\hline
					Hint (attr) & ${26.08} \pm {3.53}$  & ${31.25} \pm {4.38}$ & ${27.11} \pm {3.94}$  \\
					PKT (attr) & $\textbf{47.26} \pm \textbf{3.20}$ & $\textbf{54.94} \pm \textbf{4.18}$ & $\textbf{48.72} \pm \textbf{4.20}$  \\
					
				\end{tabular}	
	\end{minipage}%
	\begin{minipage}{.3\linewidth}
		\centering		
			\caption{VOC2007 Evaluation (object detection)}
			\label{table:voc-results}			
			\begin{tabular}{l|c}
				\textbf{Initial Model} &  \textbf{mAP} \\
				\hline
				Random & 18.39  \\
				Darknet & 38.02 \\
				PKT & \textbf{44.14}\\
			\end{tabular}
			\end{minipage} 
\end{table}

The evaluation results are reported in Table~\ref{table:sun-results}. First, the knowledge is transferred from $2 \times 2$ HoG features~\cite{dalal2005histograms,patterson2012sun} into the student network. Using the proposed PKT method leads to 26.84\% mAP outperforming the hint-based transfer (16.40\% mAP). As it was expected the student network performs slightly worse than the teacher model (HoG). This can be also attributed to the smaller representation extracted from the student network (64 dimensions instead of 300 dimensions for the HoG features). 

The proposed method was also evaluated under a cross-modal KT setup where the knowledge is transferred from the textual modality (expressed in the form of a list of textual attributes) into the student neural network that operates within the visual modality. This setup is abbreviated as ``attr'' in Table~\ref{table:sun-results}. Transferring the knowledge from the textual modality  improves the precision of the student network for both the hint-based KT method and the proposed PKT approach. Again, the proposed PKT method significantly outperforms the hint-based method leading to over 80\% relative increase of the mAP.

\textbf{KT from Object Detectors:} Finally, the proposed method was evaluated on the {PASCAL} VOC 2007 and the PASCAL VOC 2012 datasets~\cite{pascal-voc-2007,pascal-voc-2012}, following the experimental setup described in~\cite{redmon2016you}. The Darknet framework was used for training and evaluating the object detectors~\cite{redmon2016yolo9000} using the default parameters. The YOLO object detector is  usually trained after initializing its convolutional layers using another network trained for solving a  classification problem on the ImageNet dataset~\cite{krizhevsky2012imagenet}. However, with the increasing need  for more lightweight architectures this is not always possible, since usually pre-trained models exist only for a few architectures and it can be very tedious to successfully retrain every new model using such  datasets~\cite{Shen_2017_ICCV}.  The proposed PKT method can address this problem by directly transferring the knowledge from the 29-th layer (1280 filters) of the larger YOLOv2 model to the 13th layer of a smaller teacher model (a modified Tiny YOLO model with 512 filters instead of 1024 filters in its 13th layer). The proposed knowledge transfer method ran for 50 epochs with batch size 8 (due to memory constraints) and learning rate $10^{-4}$. 

The following networks were evaluated in the conducted experiments: a) a randomly initialized network (abbreviated as ``Random''), b) a network initialized using the first 11 matching layers of the Darknet reference model trained on the ImageNet dataset (abbreviated as ``Darknet'')~\cite{redmon2016yolo9000}, and c) the network trained using the proposed KT method (abbreviated as ``PKT''). The networks were trained for 40,000 iterations with batch size 64 and the results are reported in Table~\ref{table:voc-results}. First, the importance of using a knowledge transfer technique (either direct or indirect) is highlighted. Using a randomly initialized network we were unable to train a useful object detector (mAP $<$ 19\%), even though aggressive data augmentation techniques were used. This is expected, since it is well established that training deep object detectors from scratch it is not straightforward~\cite{Shen_2017_ICCV}. On the other hand, both directly using the weights of a network trained on another task and transferring the knowledge from another detector yield significantly better results. The proposed method also outperforms all the other evaluated methods leading to better object detection precision (44.14\% mAP) demonstrating  its ability to efficiently transfer the knowledge from an object detector into another one.  Note that in contrast to the baseline (``Darknet'') method, the proposed method is able to directly transfer the knowledge from a more powerful model that was trained for the task at hand (object detection) instead of using a smaller network that has been trained to extract representations for another task (classification). This can also significantly reduce the training time, since only the ``useful'' knowledge for the task at hand is transferred, instead of learning a generic object classifier using a huge dataset. 

\section{Conclusions}
\label{section:conclusions}
In this paper a Probabilistic KT technique that overcomes several limitations of existing KT methods by matching the probability distribution of the data in the feature space instead of their actual representation was proposed. %The proposed KT method is capable of embedding the manifolds formed in the feature space of the teacher model into the feature space of the student model regardless of the actual dimensionality of these spaces. 
The proposed method was extensively evaluated and it was demonstrated that outperforms several other KT techniques using different evaluation setups (KT from deep neural networks, handcrafted feature extractors, and different modalities). 

\bibliographystyle{splncs04}
\bibliography{mybib}

\end{document}